# Error Correction Maximization for Deep Image Hashing

Xiang Xu[1]
xiangxu@andrew.cmu.edu

Xiaofang Wang[2]
xiaofan2@cs.cmu.edu

Kris M. Kitani[2]
kkitani@cs.cmu.edu

[1] Department of Electrical and Computer Engineering
Carnegie Mellon University
Pittsburgh, USA

[2] The Robotics Institute
Carnegie Mellon University
Pittsburgh, USA

## Abstract

We propose to use the concept of the Hamming bound to derive the optimal criteria for learning hash codes with a deep network. In particular, when the number of binary hash codes (typically the number of image categories) and code length are known, it is possible to derive an upper bound on the minimum Hamming distance between the hash codes. This upper bound can then be used to define the loss function for learning hash codes. By encouraging the margin (minimum Hamming distance) between the hash codes of different image categories to match the upper bound, we are able to learn theoretically optimal hash codes. Our experiments show that our method significantly outperforms competing deep learning-based approaches and obtains top performance on benchmark datasets.

## 1 Introduction

Many state-of-the-art deep learning-based image hashing approaches [2, 8, 10, 11, 12, 19, 22, 23] use loss functions designed by the intuition that similar images should be close and that dissimilar images should be far in the hash code space. While these modern approaches obtain impressive performance on the task of deep image hashing, the precise theoretical definition of what 'far' should mean in terms of the Hamming distance between hash codes is still unclear. In contrast, work in coding theory provides precise definitions and theoretical bounds on the distance between optimal binary codes [14]. The goal of this work is to utilize the theoretical results of coding theory to better inform the design of loss functions for deep learning based image hashing.

The concept of the Hamming bound [14] is useful for understanding the trade-off between the length $L$ of a *codeword* (*i.e.*, a vector of $L$ binary values), the size $M$ of the *codebook* (*i.e.*, a set of $M$ unique codewords) and the minimum Hamming distance $d_{\min}$ between any two unique codewords. The Hamming bound, which is an upper bound on the codebook size, can be derived as:

$$M \leq \frac{Q^L}{f(d_{\min}, Q, L)}, \tag{1}$$

where the alphabet size $Q = 2$ for binary codes and $f$ is a function that will be described in detail later.

Typically, the Hamming bound is used to determine the maximum number of codewords $M$ that one can utilize while ensuring a desired level of error correction ability. The error correcting measure of the coding scheme depends on how different the codewords are from each other, which is directly related to the value of the minimum Hamming distance $d_{\min}$. If the codeword length $L$ and the codeword space are determined, then increasing the error correcting properties of the codewords will increase the denominator $f(d_{\min}, Q, L)$ and therefore reduces the number of codewords $M$ that can be used. We will show later that this tight coupling between codeword length, codebook size and minimum Hamming distance is very informative for the design of deep image hashing loss functions.

In the context of deep image hashing, the number of codewords (size of the codebook) $M$ is equal to the number of semantic categories in our image dataset. The intuition here is that images within a semantic category should be mapped to the same hash code and therefore we only need $M$ unique hash codes to represent $M$ classes. More importantly, $M$ is known in advance, as it is determined by the labeled image dataset. In addition, the length of the codeword $L$ (determined by the designer) and the codeword space $Q$ (typically binary) are also known prior to training the model. Given these known constants, we are able to compute an upper bound for the error correction measure or the minimum Hamming distance $d_{\min}$ as:

$$f(d_{\min}) \leq \frac{Q^L}{M}. \tag{2}$$

Here we have denoted $f(d_{\min}, Q, L)$ in shorthand as $f(d_{\min})$ since $f$ is a monotonically increasing function of $d_{\min}$ when $L$ and $Q$ are fixed. Eq. 2 describes the upper bound on the minimum Hamming distance $d_{\min}$ (more detailed description in Section 3). When one formulates the loss function for deep image hashing, it is ideal to learn a set of hash codes for which the minimum Hamming distance reaches this theoretical limit.

Based on this code theoretic perspective, we propose to directly incorporate the concept of the Hamming bound into the objective function for learning image hash codes. The key idea of our method is to define an objective function that penalizes hash codes for which the distance between two different image categories falls below the known upper bound for the minimum Hamming distance $d_{\min}$. The overall training pipeline is shown in Figure 1. Extensive experiments on standard benchmarks show that our proposed code-theoretic deep hashing approach outperforms state-of-the-art deep hashing approaches and achieve significantly better results.

## 2  Related Work

Most existing hashing methods can be categorized into unsupervised or supervised methods. Unsupervised hashing methods only utilize the training data points to learn hash codes without using any supervised information. Compared to unsupervised methods, supervised methods usually can achieve competitive performance with fewer bits due to the help of supervised information [10]. Typically, the supervised information is provided in one of three forms: pointwise labels, pairwise labels or ranking labels [12]. The representative of traditional supervised hashing methods includes CCA-ITQ [6], minimum Loss Hashing (MLH) [17], Supervised Hashing with Kernels (KSH) [15], Ranking-based Supervised Hashing(RSH) [18] and Column Generation Hashing (CGHASH) [13].

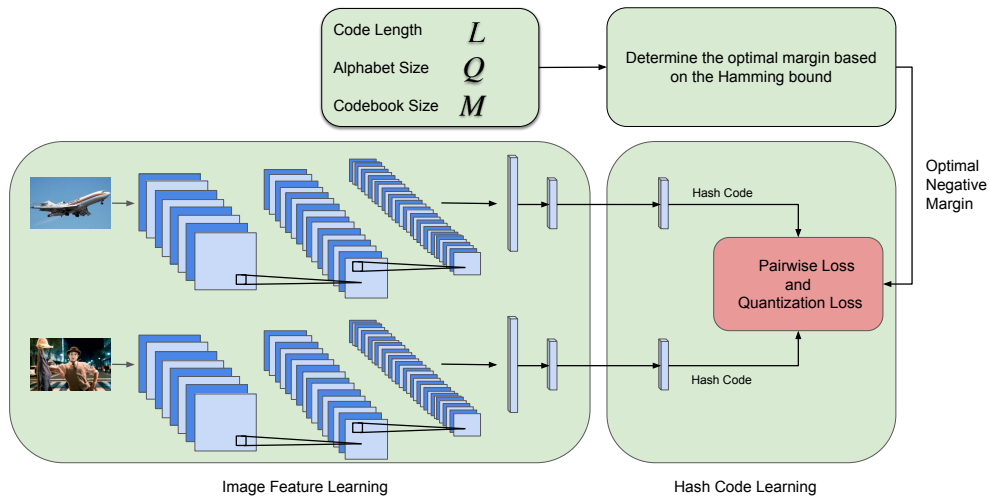

Figure 1: Overview of the proposed deep hashing method.

Recently, deep hashing methods [2, 10, 12, 19, 21, 22, 23] have been proposed to simultaneously learn image feature and hash codes with deep neural networks and have demonstrated superior performance over traditional hashing methods. One key to the success of current deep hashing methods is their carefully designed loss functions, which evaluates how well the learned hash codes approximate the similarity between images. The previous state-of-the-art method DPSH [12] proposes a definition of pairwise label likelihood to evaluate the quality of the learned hash codes and DTSH [19] further proposes a definition of triplet label likelihood. Other works such as HashNet [2] focus on designing new continuous activation function in order to address the gradient problem for training non-smooth binary hash codes.

Most loss functions are designed based on the intuition that two similar images should be close to each other and that two dissimilar images should be far from each other. However, many loss functions lack a theoretical explanation of how far two dissimilar images should be in the Hamming space. Different from them, we first show that there exists an upper bound for the minimum Hamming distance between dissimilar images and then formulate our loss function based on the bound.

## 3 Upper Bound for Minimum Hamming Distance

We are given $N$ training images $\mathcal{I} = \{I_1, \ldots, I_N\}$ from $M$ semantic classes, as well as $U$ pairwise labels. We denote the given $U$ pairwise labels as $\mathcal{S} = \{s_{q_{11}q_{12}}, \ldots, s_{q_{U1}q_{U2}}\}$, where $s_{q_{u1}q_{u2}}$ indicates whether the image of index $q_{u1}$ ($I_{q_{u1}}$) and index $q_{u2}$ ($I_{q_{u2}}$) are similar to each other or not. $s_{q_{u1}q_{u2}} = 1$ means $I_{q_{u1}}$ and $I_{q_{u2}}$ are similar to each other, and $s_{q_{u1}q_{u2}} = 0$ means $I_{q_{u1}}$ and $I_{q_{u2}}$ are dissimilar from each other. If $s_{q_{u1}q_{u2}} = 1$, we call this image pair as a positive image pair, otherwise, a negative pair.

Our goal is to learn a hash code $\mathbf{b}_n$ for each image $I_n$, where $\mathbf{b} \in \{+1, -1\}^L$ and $L$ is the target length of hash codes. Hash codes for all images $\mathcal{B} = \{\mathbf{b}_n\}_{n=1}^N$ should satisfy all the pairwise labels $\mathcal{S}$ as much as possible in the Hamming space. More specifically, given pairwise label $s_{q_{u1}q_{u2}}$, $\mathrm{dist}_H(\mathbf{b}_{q_{u1}}, \mathbf{b}_{q_{u2}})$ should be as small as possible if $s_{q_{u1}q_{u2}} = 1$, otherwise $\mathrm{dist}_H(\mathbf{b}_{q_{u1}}, \mathbf{b}_{q_{u2}})$ should be as large as possible. Here, $\mathrm{dist}_H(\cdot, \cdot)$ denotes the Hamming distance between two hash codes.

## 3.1 Hamming Bound

We now introduce the concept of *Hamming bound* from the field of coding theory [14]. Let $\mathcal{C}$ denote a codebook with a nonempty set of $Q$-ary codewords of length $L$. A $Q$-ary codeword of length $L$ refers to a $L$-length sequence $\{w_1, w_2 \ldots w_L\}$ with each element $w_i \in A$, where $A$ is the set of $Q$ possible values $\{a_1, a_2, \ldots, a_Q\}$. Each element $c \in \mathcal{C}$ is a *codeword*. We use $M$ to denote the the number of codewords in $\mathcal{C}$ and thus $M = |\mathcal{C}|$. In this work, we are only concerned with binary codes where $Q = 2$, so each codeword or hash code is a $L$-bit sequence of $-1$ and 1s.

For a codebook $\mathcal{C}$ containing at least two codewords, the *minimum Hamming distance* of this codebook, denoted by $d_{\min}(\mathcal{C})$ is,

$$d_{\min}(\mathcal{C}) = \min\{\text{dist}_H(\mathbf{x}, \mathbf{y}) : \mathbf{x}, \mathbf{y} \in \mathcal{C}, \mathbf{x} \neq \mathbf{y}\}.$$

Here $\text{dist}_H(\cdot, \cdot)$ denotes the Hamming distance between any two codewords.

To understand why the minimum Hamming distance $d_{\min}(\mathcal{C})$ is important, intuitively, we can see that the larger $d_{\min}(\mathcal{C})$ is, the more noise a code can have while still being classified correctly. Formally, if there are $V$ bits of errors in an erroneous hash code from class $m$. Then as long as $V \leq \lfloor \frac{d_{\min}(\mathcal{C})-1}{2} \rfloor$, we can simply choose the nearest codeword from $\mathcal{C}$ in terms of Hamming distance to the erroneous hash code and it will still be the same codeword from class $m$. This means that the minimum Hamming distance is a good indication of model robustness and by increasing the minimum Hamming distance, we are guaranteed to improve the image retrieval performance.

Conceptually, we hope that the codebook size $M$ and the minimum Hamming distance $d_{\min}$ are as large as possible. However, the Hamming bound shows that with fixed alphabet size $Q$, codeword length $L$ and the desired error-correcting capability (indicated by the minimum Hamming distance $d_{\min}$), there exists an upper bound for the codebook size $M$. Formally, for any $Q$-ary $(L, M, d_{\min})$-codebook, the following relationship between the codebook size $M$ and the minimum Hamming distance $d_{\min}$ must hold:

$$M \leq \frac{Q^L}{\sum_{i=0}^{\lfloor (d_{\min}-1)/2 \rfloor} \binom{L}{i}(Q-1)^i}. \tag{3}$$

The detailed proof of Hamming bound can be found in the book [14].

For the task of deep image hashing, there are $M$ semantic classes and we hope all images from one semantic class are mapped to the same hash code. Thus the goal of supervised hashing is to learn $M$ unique hash codes or a codebook of size $M$. The alphabet size $Q$ of the code is 2 as hash codes are binary values of -1 and 1s. The hash code length $L$ is also determined prior to training. We can now modify Eq. 3 as follows:

$$\sum_{i=0}^{\lfloor (d_{\min}-1)/2 \rfloor} \binom{L}{i} \leq \frac{2^L}{M}. \tag{4}$$

$\sum_{i=0}^{\lfloor (d_{\min}-1)/2 \rfloor} \binom{L}{i}$ is monotonically increasing with respect to $d_{\min}$, therefore for given $M$ and $L$ we can find the maximum $d_{\min}^*$ that satisfies this bound. This will be the minimum Hamming distance that our model optimizes towards.

# 4 Method

Our method consists of three key components: (1) image feature learning, (2) hash code learning and (3) the loss function, as shown in Figure 1. For image feature learning, we use convolutional layers and fully connected layers to learn the image feature. We initialize these layers with the weights pre-trained on ImageNet. In order to generate hash codes from image features, we adopt one fully connected layer with randomly initialized weights and set the number of nodes of the layer to the target hash code length $L$. The two components are similar to what being used in DPSH [12] and DTSH [19]. The loss function evaluates the quality of the generated hash codes and guides the training of the network. We define the loss function based on the optimal negative margin determined by the upper bound on the minimum Hamming distance. We will introduce the loss function and the optimal negative margin in detail in the following text.

## 4.1 Loss Function Definition

Our loss function measures how well a given pairwise label is satisfied by the hash codes generated by the deep neural network. The key motivation of the loss function is that it only penalizes the network when the Hamming distance between the hash codes of two dissimilar images is smaller than the upper bound on the minimum Hamming distance $d_{\min}$ computed from Eq. 4.

We use $\Theta_{ij}$ to denote the inner product between two hash codes $\mathbf{b}_i, \mathbf{b}_j \in \{+1, -1\}^L$:

$$\Theta_{ij} = \mathbf{b}_i^T \mathbf{b}_j. \tag{5}$$

Given $U_P$ as the set of indexes for positive image pairs, $U_P = \{u \mid 1 \le u \le U, s_{q_{u1}q_{u2}} = 1\}$, and $U_N$ as the indexes for negative image pairs, $U_N = \{u \mid 1 \le u \le U, s_{q_{u1}q_{u2}} = 0\}$. Our proposed pairwise label based loss function is defined as

$$\mathcal{L}_{\text{pairwise}} = \frac{1}{|U_P|} \sum_{u \in U_P} \frac{([\Theta_{q_{u1}q_{u2}} - \alpha_{\text{pos}}]_-)^2}{\alpha_{\text{pos}}^2} + \frac{1}{|U_N|} \sum_{u \in U_N} \frac{([\Theta_{q_{u1}q_{u2}} - \alpha_{\text{neg}}]_+)^2}{\alpha_{\text{neg}}^2} \tag{6}$$

where $[x]_- = \min(0, x)$, $[x]_+ = \max(0, x)$, and $\alpha_{\text{pos}}$, $\alpha_{\text{neg}}$ are the two margin parameters computed from the Hamming bound (described in Section 4.2).

We now show how minimizing $\mathcal{L}_{\text{pairwise}}$ matches the objective of pulling the hash codes of two similar images closer and pushing the hash codes of two dissimilar images further in the hash code space. We can prove the following relationship between the Hamming distance between two hash codes and their inner product:

$$\text{dist}_H(\mathbf{b}_i, \mathbf{b}_j) = \frac{1}{2}(L - \Theta_{ij}), \tag{7}$$

where $L$ is the length of the hash codes.

We first focus on the positive image pairs in Eq. 6. Assume image $I_{q_{u1}}$ and image $I_{q_{u2}}$ are similar to each other, *i.e.*, $s_{q_{u1}q_{u2}} = 1$ and they belong to the same category. Since $[x]_- = \min(0, x)$, we know that as $x$ increases, $([x]_-)^2$ will become smaller and smaller until it becomes zero. Thus, for a positive image pair $I_{q_{u1}}$ and $I_{q_{u2}}$, the larger $(\Theta_{q_{u1}q_{u2}} - \alpha_{\text{pos}})$ is, the smaller the loss $\mathcal{L}_{\text{pairwise}}$ will be. According to Eq. 7, the larger $\Theta_{q_{u1}q_{u2}}$ is, the smaller $\text{dist}_H(\mathbf{b}_{q_{u1}}, \mathbf{b}_{q_{u2}})$ will be. Therefore, by minimizing $\mathcal{L}_{\text{pairwise}}$, we can enforce the Hamming distance between two similar images to be small. Similarly, we can show that, by minimizing $\mathcal{L}_{\text{pairwise}}$, we can enforce the Hamming distance between two dissimilar images to be large.

In practice, we can also extend the loss function with the 'class-wise' strategy used in [16, 20]. The 'class-wise' strategy maintains a center hash code for each image class during training and images from other classes only need to be compared to the center hash code instead of all the images in that class. The 'class-wise' strategy can speed up the training and is useful when the number of image classes is large. More importantly, it is perfectly consistent with our derivation of the upper bound on the minimum Hamming distance $d_{\text{min}}$. In our derivation, we assume each image class is represented by a unique codeword, which is rarely true in practice. Therefore, it is reasonable to compare an image to the center hash code instead of any individual image in that class. We empirically find that 'class-wise' strategy yields better results on ImageNet-100.

Note that $\mathcal{L}_{\text{pairwise}}$ is positive only when the inner product between the hash codes of two dissimilar images is larger than $\alpha_{\text{neg}}$ or when the inner product between the hash codes of two similar images is smaller than $\alpha_{\text{pos}}$. We will introduce how we set the value of the two margin parameters based on the theoretical results derived from the Hamming bound, instead of setting them heuristically.

## 4.2 Setting the Margin

In the analysis in Sec 3, we assume that all images from the same semantic class are mapped to the same hash code. This means that the inner product between the hash codes of two similar images should equal the length of the hash code. Thus, we set $\alpha_{\text{pos}}$ to the length of the hash code, *i.e.*, $\alpha_{\text{pos}} = L$.

According to Sec 3, given the number of semantic classes $M$ and the length of the hash code $L$, we can compute the upper bound on the minimum Hamming distance between any two dissimilar images. In practice, we compute the integer value $d_{\text{min}}^*$ such that $\sum_{i=0}^{\lfloor ((d_{\text{min}}^*-1)-1)/2 \rfloor} \binom{L}{i} \leq \frac{2^L}{M}$ and $\sum_{i=0}^{\lfloor (d_{\text{min}}^*-1)/2 \rfloor} \binom{L}{i} > \frac{2^L}{M}$. Note that $d_{\text{min}}^*$ does not satisfy the Hamming bound inequality in Eq. 4 but $(d_{\text{min}}^*-1)$ does satisfy the inequality. The $d_{\text{min}}^*$ we use is one Hamming distance larger than the upper bound given by Eq. 4.

The value of $d_{\text{min}}^*$ implies that for a given $M$ and $L$, the minimum Hamming distance between dissimilar images cannot be larger than $d_{\text{min}}^*$. Therefore, we set the value of the margin $\alpha_{\text{neg}}$ such that the loss function only penalizes the network when the Hamming distance of two dissimilar images is smaller than $d_{\text{min}}^*$. According to Eq. 7, we can compute the value of $\alpha_{\text{neg}}$ using the following equation:

$$\alpha_{\text{neg}} = L - 2d_{\text{min}}^*. \tag{8}$$

Note that here we set the value of $\alpha_{\text{neg}}$ based on the theoretical results derived from the Hamming bound. This stands in sharp contrast to previous work where the margin parameter is set heuristically.

## 4.3 Quantization Loss

We denote the vector of activations of the last layer of the network for the $n$-th image as $\mathbf{u}_n$ and we obtain the binary codes $\mathbf{b}_n$ by applying the sign function to $\mathbf{u}_n$, *i.e.*, $\mathbf{b}_n = sgn(\mathbf{u}_n)$. Directly minimizing $\mathcal{L}_{\text{pairwise}}$ with back-propagation is impossible since the gradient of the sign function is always 0. Therefore, during training, we relax the binary codes $\mathbf{b}_n$ to the float vector $\mathbf{u}_n$ and re-define $\Theta_{ij}$ as $\Theta_{ij} = \mathbf{u}_i^T \mathbf{u}_j$. Using this surrogate loss function, we can minimize $\mathcal{L}_{\text{pairwise}}$ with back-propagation. Since the approximation will introduce additional quantization error, we also include the quantization error term in our loss function, following DTSH [19] and DPSH [12]. Our final loss function becomes,

$$\mathcal{L}_{\text{total}} = \mathcal{L}_{\text{pairwise}} + \lambda \mathcal{L}_{\text{quan}}, \tag{9}$$

where $\mathcal{L}_{\text{quan}}$ is defined as $\mathcal{L}_{\text{quan}} = \sum_{n=1}^{N} ||\mathbf{b}_n - \mathbf{u}_n||_2^2$ and $\lambda$ is the hyper-parameter to balance $\mathcal{L}_{\text{pairwise}}$ and $\mathcal{L}_{\text{quan}}$.

# 5 Experiments

## 5.1 Datasets and Settings

We conduct experiments on two standard benchmarks: CIFAR-10 [9] and ImageNet-100 [4]. Each dataset is split into a database image set and a test image set. Training images and validation images are sampled from the database image set. At test time, images in the test set are used as query images to query the database. The retrieval performance is evaluated using Mean Average Precision (MAP) for experiments conducted on CIFAR-10. For ImageNet-100, we use MAP@1000 as the evaluation metric,

**CIFAR-10** contains 60K images of size 32×32. It has 10 different categories and 6,000 images for each category. Following the experiment setup in [10, 12, 19, 21], we evaluate our method under two different settings. In the first setting, 1K images (100 images per class) are randomly selected as the test set. The remaining images are used as the database. 5K database images (500 images per class) are randomly sampled as the training set and 1K database images (100 images per class) are randomly sampled as the validation set. In the second setting, 10K images (1K images per class) are randomly sampled as test images. The remaining 50K images are used as database images and all the database images are used for training.

**ImageNet-100** is a subset of ImageNet [4] with 100 randomly sampled classes. We use the same data split as HashNet [2]. All images from ILSVRC 2012 train set are treated as the database images, and 130 images per class, totally 130K images, are randomly sampled from the database images as training data. All images in the selected classes from the ILSVRC 2012 validation set are used as test images. 50 images per class are randomly sampled from the database as the validation set.

Following the setup from previous works [1, 2, 12, 19], we use the pre-trained VGG-F network [3] to initialize our network for experiments on CIFAR-10. Our network is trained with SGD and an initial learning rate of 0.05. For experiments on ImageNet-100, we use pre-trained AlexNet [9] and set the learning rate for fully-connected layers to be 10 times larger than that of the convolutional layers, similar to HashNet [2]. The initial learning rate is set to 0.005 and the model is trained using SGD with 0.5 momentum. The batch size is set to 64 for all the experiments.

## 5.2 Performance Evaluation

We have tried training the network with the loss function in Eq. 9 (denoted by 'Ours') and the loss function extened with the 'class-wise' strategy (denoted by 'Ours + class-wise').

| Method | CIFAR-10 (5K train / 1K test) | | | |
|---|---|---|---|---|
| | bits = | 12 | 24 | 32 | 48 |
| | $\alpha_{neg} =$ | -6 | -14 | -18 | -34 |
| MIHash [5] | | 0.687 | 0.775 | 0.786 | 0.822 |
| TALR-AP [7] | | 0.732 | 0.789 | 0.800 | 0.826 |
| DPSH [12] | | 0.720 | 0.757 | 0.757 | 0.767 |
| DTSH [19] | | 0.725 | 0.773 | 0.781 | 0.810 |
| **Ours** | | **0.832** | **0.858** | **0.864** | **0.870** |
| **Ours + class-wise** | | **0.835** | **0.854** | **0.862** | **0.866** |

Table 1: MAP on CIFAR-10 (5K train / 1K test).

| Method | CIFAR-10 (50 K train / 10K test) | | | |
|---|---|---|---|---|
| | bits = | 16 | 24 | 32 | 48 |
| | $\alpha_{neg} =$ | -6 | -14 | -18 | -34 |
| MIHash [5] | | 0.929 | 0.933 | 0.938 | 0.942 |
| TALR-AP [7] | | 0.939 | 0.941 | 0.943 | 0.945 |
| DPSH [12] | | 0.908 | 0.909 | 0.917 | 0.932 |
| DTSH [19] | | 0.916 | 0.924 | 0.927 | 0.934 |
| **Ours** | | **0.950** | **0.951** | **0.953** | **0.952** |
| **Ours + class-wise** | | **0.949** | **0.950** | **0.952** | **0.952** |

Table 2: MAP on CIFAR-10 (50K train / 10K test).

**CIFAR-10**: We set $\lambda$ to 0.002 for all experiments on CIFAR-10. The Mean Average Precision (MAP) for the two settings are shown in Table. 1 and Table. 2 respectively. 'Ours + class-wise' refers to the performance when using the 'class-wise' strategy. The values of the negative margin used in our experiments are also shown in the table. They are determined based on Eq. 8. Experimental results from both settings show that our method outperforms previous deep hashing methods across different hash code lengths, whether using the 'class-wise' strategy or not.

**ImageNet-100**: To further validate our proposed method, we conduct experiments on the more challenging ImageNet-100 dataset. ImageNet-100 has more classes then CIFAR-10 and images are more diverse, which makes it harder to learn high-quality hash codes. For all thee experiments on ImageNet-100, we set $\lambda$ to 0.01. Table 3 summarizes the MAP@1K for different code lengths. We see that our method performs on par with the previous state-of-the-art method and significantly outperforms them when combined with 'class-wise' strategy. On ImageNet-100, the best performance is achieved when extending our method with the 'class-wise' strategy. We attribute the performance gain to that using the 'class-wise' strategy is more consistent with the derivation of the upper bound as explained in Sec. 4.1.

## 5.3　Ablation Studies

**Impact of Negative Margin**: We now study the impact of negative margin $\alpha_{neg}$ on the final performance. All other settings except for the negative margin $\alpha_{neg}$ are unchanged. All the following experiments are conducted with the 'class-wise strategy'.

Figure. 2(a) summarizes the results for CIFAR-10 with different negative margin values under the first setting. Performance of the negative margin computed based on the Hamming bound is shown in black dots. We can see that the negative margin determined by the Hamming bound consistently give the best performance or one of the best performances across different code lengths. The performance is more sensitive to the negative margin value for shorter hash codes than longer hash codes. Shorter hash codes provide a smaller space for all the classes than long hash codes, making it harder for the network to learn to generate high-quality hash codes, thus setting the negative margin to an inappropriate value will lead to a larger drop in terms of performance.

| Method | bit= | 16 | 32 | 48 | 64 bits |
|---|---|---|---|---|---|
| | $\alpha_{neg}=$ | 2 | -6 | -18 | -30 |
| HashNet [2] | | 0.5059 | 0.6306 | 0.6633 | 0.6835 |
| MIHash [5] | | 0.5688 | 0.6608 | 0.6852 | 0.6947 |
| TALR-AP [7] | | 0.5892 | 0.6689 | 0.6985 | 0.7053 |
| **Ours** | | **0.6032** | 0.6658 | 0.6892 | 0.7014 |
| **Ours + class-wise** | | **0.6429** | **0.6967** | **0.7163** | **0.7247** |

Table 3: MAP@1000 on ImageNet-100.

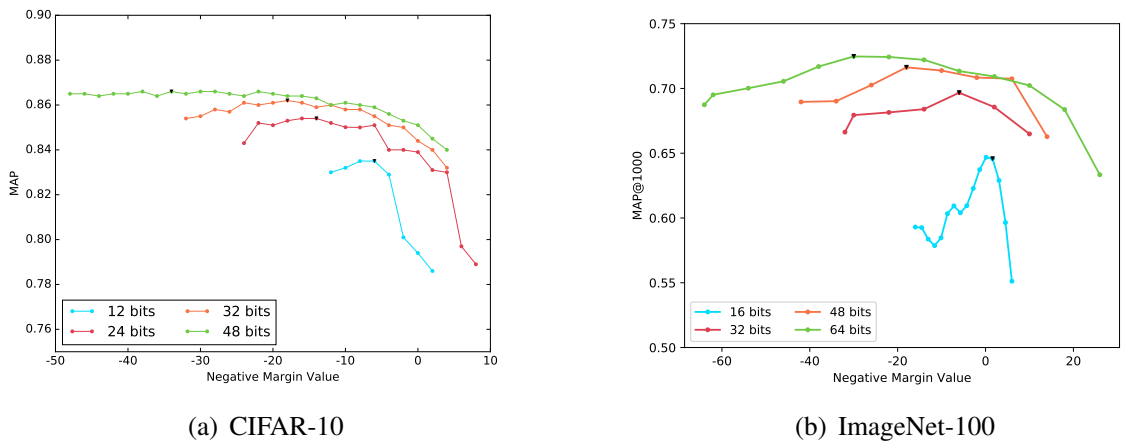

(a) CIFAR-10          (b) ImageNet-100

Figure 2: Impact of negative margin. Black dots represents the negative margin computed based on the Hamming bound.

We also show the results on ImageNet-100 in Figure 2(b). We can see that on ImageNet-100, the negative margin determined by the Hamming bound also consistently gives the best performance. We also observe that the performance is more sensitive to the negative margin value on ImageNet-100 than that on CIFAR-10. This is because ImageNet-100 has much more classes and also images in ImageNet-100 are more diverse. Setting the negative margin to an inappropriate value will make it harder to represent all the images as hash codes while still preserving the similarity between them.

**Impact of Hyper-Parameter**: We now study the impact of the hyper-parameter of $\lambda$ on the performance. We vary the value of $\lambda$ and report the MAP@1000 on ImageNet-100 with code length 32 and 64. The results are shown in Figure 3. We can see that our method works well when $\lambda$ is between 0.001 and 0.1. When $\lambda = 1.0$, we observe that the training does not converge so we exclude those points from the figure.

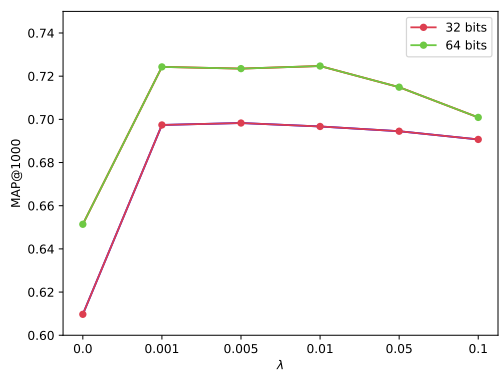

Figure 3: Impact of $\lambda$ on ImageNet-100.

# 6   Conclusion

In this paper, we have proposed a principled method for designing the loss function of a deep image hashing network. We show that the minimum Hamming distance between two images from different semantic categories has an analytic upper bound and that by incorporating that value into the loss function, we can obtain an optimal hash code. As a consequence, we not only improve the error correction ability of the learned hash codes, but moreover, we showed through empirical validation that our upper bound based loss function leads to significant improvements in the quality of the generated hash codes over state-of-the-art deep image hashing approaches.